\documentclass[sigconf,nonacm]{acmart}

\begin{document}
\title{\vspace{-1.45em}Evolving Neuronal Plasticity Rules \\ using Cartesian Genetic Programming}
\date{\today}

\author{Henrik D. Mettler}
\affiliation{
  \institution{\small Department of Physiology, University of Bern}
}
\authornote{Correspondence: henrik.mettler@unibe.ch}

\author{Maximilian Schmidt}
\affiliation{
  \institution{\small RIKEN Center for Brain Science, Tokyo, Japan}
}

\author{Walter Senn}
\affiliation{
  \institution{\small Department of Physiology, University of Bern}
}

\author{Mihai A. Petrovici}
\affiliation{
  \institution{\small Department of Physiology, University of Bern \\
  Kirchhoff Institute for Physics, Heidelberg University}
 }
  
\authornote{ Shared senior authorship}

\author{Jakob Jordan}
\affiliation{
  \institution{\small Department of Physiology, University of Bern}
  }
\authornotemark[2]

\renewcommand{\shortauthors}{H.~D.~Mettler et al.}

\begin{abstract}
We formulate the search for phenomenological models of synaptic plasticity as an optimization problem.
We employ Cartesian genetic programming to evolve biologically plausible human-interpretable plasticity rules that allow a given network to successfully solve tasks from specific task families. 
While our evolving-to-learn approach can be applied to various learning paradigms, here we illustrate its power by evolving plasticity rules that allow a network to efficiently determine the first principal component of its input distribution.
We demonstrate that the evolved rules perfom competitively with known hand-designed solutions.
We explore how the statistical properties of the datasets used during the evolutionary search influences the form of the plasticity rules and discover new rules which are adapted to the structure of the corresponding datasets.
\end{abstract}


\maketitle

\section{Introduction}

Changes in coupling strength between neurons in the central nervous system are believed to be central for the acquisition of new skills and memories in humans and other animals.
While the microscopic biochemical processes are extraordinarily complex, phenomenological models which describe changes in the postsynaptic response to presynaptic activity have long been explored and successfully related to experimental data \cite{BiPoo}.
Furthermore, modern approaches often provide a normative view on neuron and synapse dynamics \cite{lillicrap2020backpropagation}.
Despite these successes, the construction of new phenomenological models remains a laborious, manual process.
Here we pursue an automated approach to constructing phenomenological models of synaptic plasticity by employing genetic programming to evolve rules for synaptic that learn efficiently. 
We refer to this approach as ``Evolving to learn'' (E2L).

A simple, but useful abstraction of information processing in cortical neurons is obtained by describing a neuron's output $y_i$ as a linear, weighted sum of presynaptic activities $x_j$, followed by the application of an activation function $\rho$: $y_i = \rho \left(\sum_{j=1}^n w_{ij} x_j\right)$.
We consider plasticity rules $f$ that determine changes in the coupling strength $w_{ij}$ from neuron $j$ to $i$: $\Delta w_{ij} \propto f(X_{ij})$.
Here $X_{ij}$ represents a set of local variables, such as pre- and postsynaptic activity traces or synaptic weights.
\nobreak{We formulate the search for synaptic plasticity rules as an optimization problem \cite{bengio1992optimization}:
\begin{align}
  f^* = \text{argmax}_f \mathcal{F}(f, \Omega) \; .
\end{align}
}
Here $\mathcal{F}$ represents the fitness of rule $f$, and $\Omega$ represents the specific experimental conditions, for example the network model and task family.
The fitness measures how well a given network with plasticity rule $f$ solves tasks from the considered task family.

Recent work has defined $f$ as parametric function, using evolutionary strategies to optimize parameter values \cite{Confavreux2020}.
While this approach allows the use continuous optimization methods, the choice of the parametric form severely constraints the search space.
Other authors have encoded plasticity rules using artificial neural networks \cite{Risi2010}.
While this allows the plasticity rule to take, in principle, any computable form, the macroscopic computation by ANNs is notoriously difficult to understand, limiting the interpretability of the discovered rules.
In contrast, we aim to discover interpretable synaptic plasticity rules in large search spaces.
We employ Cartesian genetic programming (CGP) \cite{MillerCGP2019} to represent and evolve plasticity rules as compact symbolic expressions.
Previous work has successfully demonstrated this approach on various learning paradigms for spiking neuronal networks \cite{jordan2020evolving}. 
Here we explore the application to rate-based models.
As an example, we aim to discover plasticity rules that extract the first principal component of an input data set. 
We use the hand-designed ``Oja's rule'' \cite{oja1982} as a competitive baseline.

\vspace{-0.15em}
\section{Results}

The neuronal network consists of $n_\text{input}$ input units and a single output unit.
Like previous work \cite{oja1982} we consider linear activation functions $\rho(x) = x$, hence $y=\sum_{j=1}^n w_j x_j$.
A task is defined by a set $\mathcal{D}$ of $M$ input vectors $\textbf{x}$ sampled from a multi-dimensional Gaussian with zero mean and covariance matrix $\Sigma$.
In every trial $i$ we sample (without replacement) an input vector $\textbf{x}^{(i)}$ from $\mathcal{D}$, compute the output activity $y$ and update synaptic weights elementwise according to $f$: $\Delta w_j^{(i)} = \eta\, f(y^{(i)}, x_j^{(i)}, w_j^{(i-1)})$, where $\eta$ is a fixed learning rate.
Our goal is to discover rules which align the synaptic weight vector $\textbf{w}$ with the first principal component of the dataset ($\textbf{PC}_0$).
The set of all possible covariance matrices $\{ \Sigma\}$ defines a task family $\mathcal{T}_0$.
We further consider two additional task families: $\mathcal{T}_1$, where the components of $\textbf{PC}_0$ are of approximately equal amplitude and $\mathcal{T}_2$, where $\textbf{PC}_0$ is aligned with one of the axes.
We define the fitness of a plasticity rule $f$ for a dataset $\mathcal{D}$ as a sum of two terms, measuring the deviation of the weight vector from $\textbf{PC}_0$, and a regularizer for its length, respectively, averaged over $M$ trials:
\vspace{-0.15em}
\begin{equation}
  \label{eq:fit2}
   \mathcal{F}(f,\mathcal{D}) = \frac{1}{M} \sum_{i=1}^M \big| \cos{\left(\angle(\textbf{w}_i, \textbf{PC}_0)\right)} \big|- \alpha \big| ||\textbf{w}_i||_2-1 \big| \;.
\end{equation}
Here $\angle(\cdot, \cdot)$ denotes the angle between two vectors, and $\alpha > 0$ is a hyperparameter controlling the strength of the regularizer.
To avoid overfitting plasticity rules to a single dataset, we define the fitness of a plasticity rule $f$ for a task family $\mathcal{T}$ as the sampled average over $K$ datasets from this family: $\mathcal{F}(f) = \mathbb{E}_{\mathcal{T}}[\mathcal{F}(f,\mathcal{D})]$. 

\begin{figure}[tbp!]
	\includegraphics[width=0.45 \textwidth]{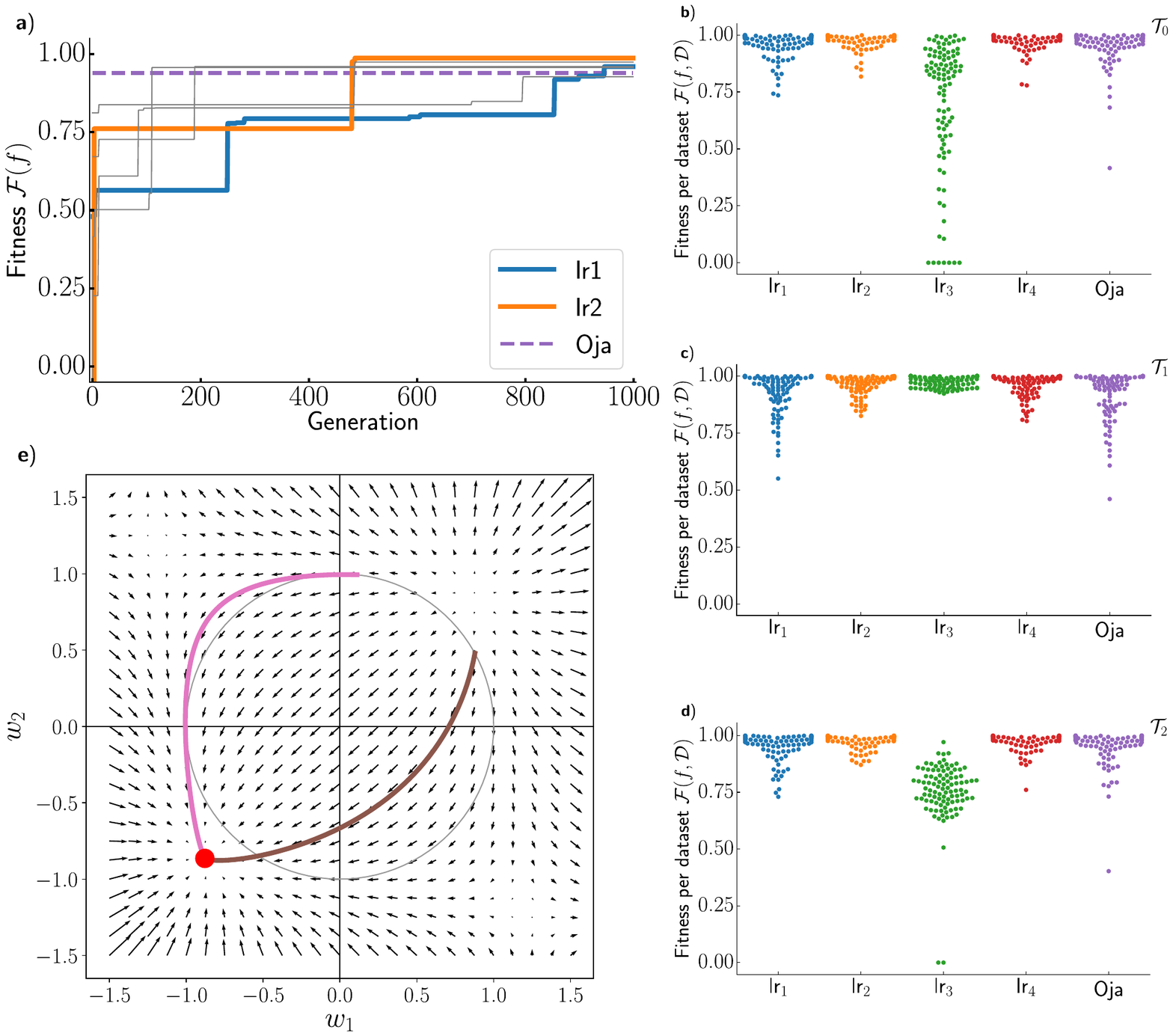}
    \vspace{-14em}

    \caption{E2L discovers plasticity rules which perform PCA.
    \textbf{a)} Fitness of the best-performing individual over generations for multiple evolutionary runs with different initial conditions with covariance matrix $\Sigma$ sampled from $\mathcal{T}_0$.
     Random initial weights for each dataset, constant across generations to make individuals from different generations comparable.
      \textbf{b-d)} Fitness per dataset for $n=100$ datasets not used in evolutionary run, with covariance $\Sigma$ sampled from $\mathcal{T}_0$ (\textbf{b)}, $\mathcal{T}_1$ (\textbf{c)} and $\mathcal{T}_2$ (\textbf{d)}.
      Parameters: $n=2, K=10, M=1000, \eta=0.01$.
      $f$ is constructed from the operator set $\{+, -, *\}$, with the input set $X_{ij} =  \left(\{w_{ij}, x_j, y\}\right)$.
      For implementation details see \cite{schmidt_maximilian_2020_3889163}.
      e) Phase plane of $\text{lr}_3$, trained on dataset, $\text{Var}[x_1]= 1.0, \text{Var}[x_2]=0.9, \text{Cov}[x_1x_2]=0.3$, with two sample tractories converging to the fixed point.
      Gray indicates possible initial weights.
      }\label{fig:Cgp_pca}
	\vspace{-1em}
\end{figure}

When trained with tasks sampled from $\mathcal{T}_0$, $5$ out of $6$ evolutionary runs with different initial conditions evolve plasticity rules which allow the network to approximate $\textbf{PC}_0$ of the respective dataset as good as or even slightly better than Oja's rule (Fig.~\ref{fig:Cgp_pca}a, b; $\Delta w_j^\text{Oja}=\eta y(x_j-w_jy), \Delta w_j^{\text{lr}_1}=\eta(2y+1+w_j)(x_j-w_jy),  \Delta w_j^{\text{lr}_2}=\eta 2y(x_j-w_jy)$).
These learning rules typically contain Oja's rule as a subexpression.
Similarly to Oja's rule, learning rules evolved on datasets with random principle components generalize well to datasets with statistical structure (Fig.~\ref{fig:Cgp_pca}c,d).
$\text{lr}_2$ slighly outperforms Oja across the investigated datasets due to a constant scaling factor which effectively increases its learning rate.
These results demonstrate that our approach is able to robustly recover efficient hand-designed plasticity rules. 

When evolved on structured data (task families $\mathcal{T}_1, \mathcal{T}_2$), learning rules tend to specialize and outperform their more general counterparts (Fig.1c, $\Delta w_j^{\text{lr}_3}=\eta(-x_j)(x_j-w_jy)$; Fig.1d, $\Delta w_j^{\text{lr}_4}=\eta(y+w_jx_j)(x_j-w_jy) $).
However, evolved rules vary in their generalizability. 
For example, $\text{lr}_3$ rule does not generalize well to datasets with different statistical structure.
The availability of plasticity rules as closed-form expressions helps us understand why. It is straightforward to derive the expected weight changes under $\text{lr}_3$ as $\mathbb{E}_{\mathcal{D}} \left[ \Delta w_j^{\text{lr}_3} \right] = \eta\left((w_j^2 - 1) \text{Var}[x_j] + w_j \sum_{i\neq j} w_i \text{Cov}[x_i x_j]\right)$.
In two dimensions, this system of equations has only one stable fixed point with a wide basin of attraction that fully covers our assumed initialization space ($||\textbf{w}||_2=1$)(Fig.~\ref{fig:Cgp_pca}e). 
For $\mathcal{D}$ from $\mathcal{T}_1$, the fixed point is close to $(-1, -1)$, thus approximately maximizing the fitness.
For $\mathcal{D}$ from $\mathcal{T}_2$, the fixed point remains close to the diagonal, which is no longer aligned with $\textbf{PC}_0$, thus prohibiting high fitness values (green dots in  Fig.~\ref{fig:Cgp_pca}c, d).
In contrast, learning rules evolved on datasets from $\mathcal{T}_2$, perform well on tasks from all task families (Fig.~\ref{fig:Cgp_pca}b,c,d), similar to Oja's rule.

\section{Conclusion}

We demonstrated that E2L can successfully discover interpretable biophysically plausible plasticity rules allowing a neuronal network to solve a well-defined task.
Not only did we recover Oja's rule, but by evolving rules on datasets with specific structure we obtained variations which are adapted to the corresponding task families.

This adaptation can be viewed as an example of ``overfitting'' that should be avoided.
However, we believe this to be an important feature of our approach: Evolving to learn from data with specific statistical structure and thus embedding empirical priors into plasticity rules could potentially explain some of the fascinating aspects of few-shot learning and quick adaptation to novel situations displayed by biological agents.
For example, it seems reasonable to expect that plasticity mechanisms driving the organization of sensory cortices are adapted to the statistical structure of their inputs, reflecting an evolutionary specialization to the ecological niche of organisms. 

\begin{acks}
This research has received funding from the European Union Horizon 2020 Framework Programme for Research and Innovation under the Specific Grant Agreement No. 945539 (Human Brain Project SGA3).
\end{acks}

\bibliographystyle{ACM-Reference-Format}
\bibliography{biblist_arxiv}

\end{document}